\documentclass[letterpaper, 10 pt, conference]{ieeeconf}
\IEEEoverridecommandlockouts

\usepackage{amsmath}
\usepackage{amssymb}
\usepackage[nospace]{cite}
\usepackage{url}
\usepackage{xcolor}
\usepackage{fancyhdr}
\usepackage{adjustbox}
\usepackage{graphicx}
\usepackage{tensor}
\usepackage{tabularx}
\usepackage{multirow}
\usepackage{multicol}
\usepackage{colortbl}
\usepackage[utf8]{inputenc}
\usepackage[OT1]{fontenc}
\usepackage[final]{microtype}
\usepackage{booktabs}
\usepackage[binary-units]{siunitx}

\usepackage[font=small,labelfont=bf]{caption}
\usepackage{cuted}

\newcommand\mytxcellwidth{\TX@col@width}

\makeatletter
\let\NAT@parse\undefined
\makeatother
\usepackage[hidelinks]{hyperref}
\hypersetup{
    colorlinks=true,     
    linkcolor=blue,      
    urlcolor=blue,       
    citecolor=blue,      
}

\newcolumntype{Y}{>{\centering\arraybackslash}X}

\usepackage{misc/onimage}

\tikzset{
    image label/.style={
        every node/.style={
            fill=black,
            text=white,
            font=\fontfamily{phv}\selectfont\scriptsize\bfseries,
            anchor=north west,
            xshift=0.05cm,
            yshift=-0.15cm,
            at={(0,1)}
        }
    }
}

\tikzset{
    image label_tl/.style={
        every node/.style={
            fill=black,
            text=white,
            font=\fontfamily{phv}\selectfont\scriptsize\bfseries,
            anchor=north west,
            xshift=0.1cm,
            yshift=-0.1cm,
            at={(0,1)}
        }
    }
}

\tikzset{
    image label_tr/.style={
        every node/.style={
            fill=black,
            text=white,
            font=\fontfamily{phv}\selectfont\scriptsize\bfseries,
            anchor=north east,
            xshift=-0.1cm,
            yshift=-0.1cm,
            at={(1,1)}
        }
    }
}

\tikzset{
    image label_bl/.style={
        every node/.style={
            fill=black,
            text=white,
            font=\fontfamily{phv}\selectfont\scriptsize\bfseries,
            anchor=south west,
            xshift=0.1cm,
            yshift=0.1cm,
            at={(0,0)}
        }
    }
}

\tikzset{
    image label_br/.style={
        every node/.style={
            fill=black,
            text=white,
            font=\fontfamily{phv}\selectfont\scriptsize\bfseries,
            anchor=south east,
            xshift=-0.1cm,
            yshift=0.1cm,
            at={(1,0)}
        }
    }
}

\title{
\vspace{1pt}
\LARGE \bf 
ViPlanner: Visual Semantic Imperative Learning for Local Navigation
\vspace{-8pt}
}

\author{
Pascal Roth$^{1}$, Julian Nubert$^{1}$, Fan Yang$^{1}$, Mayank Mittal$^{1,2}$, and Marco Hutter$^{1}$
\thanks{This work is supported in part by the Max Planck ETH Center for Learning Systems, the EU Horizon 2020 programme grant agreement No.852044, 10107045 and 101016970, the EU Horizon Europe Framework Programme grant agreement No. 101070405 and 101070596, the NCCR digital fabrication and robotics, and the SNSF project No.188596.}
\thanks{$^{1}$All authors are with the Robotic Systems Lab, ETH Z\"urich, 8092 Z\"{u}rich, Switzerland. Contact: {\tt\small\{rothpa, nubertj, fanyang1, mmittal, mahutter\}@ethz.ch}.
}    \thanks{
$^{2}$The author is with NVIDIA.}%
}

\usepackage{fancyhdr}
\newcommand{\mytitle}{\textbf{Accepted version.} To appear in the proceedings of the \textit{International Conference on Robotics and Automation (ICRA), 2024.}
} 
\begin{document}

\bstctlcite{IEEEexample:BSTcontrol}

\makeatletter
    \let\@oldmaketitle\@maketitle
    \renewcommand{\@maketitle}{
    \@oldmaketitle
    \centering
    \vspace{-2.5pt}
    \includegraphics[width=1.0\textwidth]{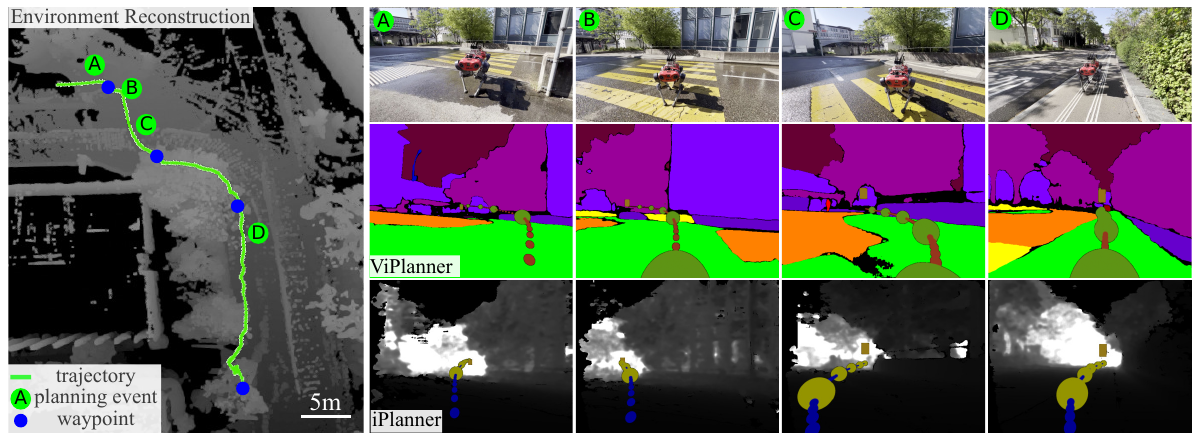}
    \captionof{figure}{Quadrupedal navigation in large scale urban environments requires semantic understanding to successfully follow side- and crosswalks. Four local planning events \mbox{(A - D)} along the autonomously traversed path are shown. The planned path of the proposed semantic imperative planner is projected into the semantic images (middle row), whereas the estimated path of the purely geometric iPlanner~\cite{yang_iplanner_2023} is overlaid onto the depth image (bottom row). The traversed path is shown in an environment reconstruction generated by~\cite{open3d_slam_2022}.}
    \label{fig:real_world_crosswalk}
    \vspace{-9.5pt}
    \addtocounter{figure}{-1}%
    }
\makeatother

\maketitle

\addtocounter{figure}{1}%

\thispagestyle{fancy}


\begin{abstract}
Real-time path planning in outdoor environments still challenges modern robotic systems due to differences in terrain traversability, diverse obstacles, and the necessity for fast decision-making. Established approaches have primarily focused on geometric navigation solutions, which work well for structured geometric obstacles but have limitations regarding the semantic interpretation of different terrain types and their affordances. Moreover, these methods fail to identify traversable geometric occurrences, such as stairs.
To overcome these issues, we introduce ViPlanner, a learned local path planning approach that generates local plans based on geometric and semantic information. The system is trained using the Imperative Learning paradigm, for which the network weights are optimized end-to-end based on the planning task objective. This optimization uses a differentiable formulation of a semantic costmap, which enables the planner to distinguish between the traversability of different terrains and accurately identify obstacles.
The semantic information is represented in 30 classes using an RGB colorspace that can effectively encode the multiple levels of traversability. We show that the planner can adapt to diverse real-world environments without requiring any real-world training. In fact, the planner is trained purely in simulation, enabling a highly scalable training data generation.
Experimental results demonstrate resistance to noise, zero-shot sim-to-real transfer, and a decrease of $38.02$\% in terms of traversability cost compared to purely geometric-based approaches. Code and models are made publicly available: \url{https://github.com/leggedrobotics/viplanner}.
\end{abstract}


\section{Introduction}

Path planning is a fundamental problem in robotics. Significant progress has been made for mobile navigation in environments where pre-built high-definition maps are available~\cite{bao_2023}. 
In contrast, several challenges still exist for planning in unknown environments fully relying on onboard sensors due to sensor noise, dynamic or moving objects, and diverse scenes~\cite{yang_iplanner_2023}. 
At present, most works address these challenges using purely geometric navigation solutions~\cite{yang_iplanner_2023,cao2022autonomous,wellhausen_rough_2021,yang_real-time_2021,loquercio_learning_2021, hoeller2021learning}. 
While showing good performance in structured or unpopulated (known) environments, such as indoors~\cite{el2021indoor} or underground environments~\cite{tranzatto2022cerberus}, navigation becomes much harder once robots enter unstructured outdoor environments.
On the one hand, upcoming systems must be able to distinguish terrain of varying traversability with the same geometric appearance (e.g., mud vs. concrete), while on the other hand, seemingly geometric obstacles (such as steps or stairs) must be interpreted correctly.
Including the semantic domain in the traversability estimation has the potential to improve the assessment in semi-structured environments~\cite{maturana2018realTime,shaban_semantic_2022,muller_driving_2018}.

For unknown environments, current path planning algorithms are either designed as end-to-end learned or modular approaches. 
In the latter, a perception module typically estimates the traversability while the path is generated by a sampling or optimization-based method~\cite{cao2022autonomous,wellhausen_rough_2021,yang_real-time_2021}. 
While these methods can generalize well, they require collecting a vast amount of difficult-to-acquire real-world data for the traversability assessment and searching a path in this map, which can introduce large latencies.
On the other hand, in end-to-end learned solutions, the path is directly predicted from sensor measurements, which reduces latencies. 
These methods are trained either through imitation learning (IL) with expert demonstrations~\cite{loquercio_learning_2021, hecker2020learning}, with reinforcement learning (RL)~\cite{hoeller2021learning, liu2020rl} or very recently via imperative learning (ImpL)~\cite{yang_iplanner_2023}. 
While IL historically suffers from low generalization capabilities due to the limited availability of demonstrations, RL and ImpL can be trained entirely in simulation or with a mix of simulated and real-world data. 
ImpL employs an offline Bi-Level Optimization (BLO) over a predefined differentiable cost (map) to generate smooth (path) predictions. 
It enhances training efficiency compared to RL and has been shown to outperform previous methods~\cite{yang_iplanner_2023}.
Nevertheless, the existing ImpL planner, called iPlanner~\cite{yang_iplanner_2023}), is restricted to the geometric domain and requires diverse real-world data to be applied safely. 

In this work, we present ViPlanner, an end-to-end learned, multi-domain local planner that uses the ImpL paradigm by building up on iPlanner. 
Our core contributions are:
\begin{enumerate}
    \item The development of a semantically-aware local planner using an unsupervised Imperative Learning approach.
    \item The achievement of zero-shot transfer from simulation to real-world by combining and integrating a pre-trained semantic segmentation network and geometric input during end-to-end training.
    \item Evaluations and benchmarks of the proposed method against the geometric-based approach~\cite{yang_iplanner_2023} in both simulated and real-world settings using the quadrupedal robot ANYmal~\cite{hutter2017anymal}.
    \item The released open-source code featuring an efficient, scalable pipeline for data generation and planner evaluation, applicable to indoor and outdoor environments, using high-fidelity simulation~\cite{mittal2023orbit} based on \textit{NVIDIA Omniverse}.
\end{enumerate}


\section{Related Work}
Local path planning has been extensively explored over the past two decades. Traditionally, a modular approach consisting of \textit{i)} a traversability estimation module, and \textit{ii)} a path-searching algorithm is applied~\cite{paden2016survey}. 
Early works assumed a mostly observable environment with given traversability estimation and tackled the local path planning problem with optimization-based~\cite{ratliff2009chomp}, sampling-based~\cite{karaman2011sampling, kavraki1996prm}, as well as heuristic/primitive-based methods~\cite{schaal2003control, dharmadhikari2020primitives}. 
Over the past few years, the techniques have gotten more advanced and can master more complex environments~\cite{wellhausen_rough_2021} or robot configurations~\cite{jelavic2023lstp}, recently often combining sampling-based and optimization-based methods in one approach~\cite{jelavic2023lstp, sleiman2023versatile}. However, the conceptual separation in \textit{i)} and \textit{ii)} has mostly remained unchanged.
Accelerated by the advent of deep learning, recent advances in literature have explored data-driven approaches, from learning-based traversability assessment~\cite{frey2023fast, wellhausen_where_2019},
up to fully end-to-end learned approaches~\cite{yang_iplanner_2023, hoeller2021learning}.

\paragraph{Modular Geometric Approaches}
Classical approaches analyze the environment primarily through geometric information in the form of point clouds~\cite{cao2022autonomous, frey2022loco} or meshes~\cite{hudson2022field} and determine traversability based on metrics such as occupancy or stepping difficulty~\cite{fan2021step}. A sampling or optimization-based path-searching module then determines the path by employing traditional techniques such as $\text{PRM}^{\ast}$~\cite{wellhausen_rough_2021}, visibility graphs~\cite{cao2022autonomous} or first learning-based approaches~\cite{yang_real-time_2021}. However, only using geometric measurements results in the failure to reason about paths over different terrains with similar geometry or predict paths that could traverse geometric obstacles, like stairs, making the planner inadequate for outdoor environments. Instead, this work focuses on a multi-domain approach for enhanced environmental understanding. 

\paragraph{Modular Semantic Approaches}
Semantics as additional domains within the traversability assessment allow for enhanced reasoning about the environment where each semantic class is assigned a traversability cost. By annotating a geometric point cloud with semantic labels, \cite{maturana2018realTime, shaban_semantic_2022} created a dense traversability map and enabled autonomous off-road navigation. Moreover, \cite{bartolomei2020perception, Vasilopoulos2020Reactive, cai2022risk} uses semantic information to improve reactiveness and the planner's safety. The resulting composition of three modules introduces large latencies in the systems and limits the application to mostly static environments. While we also include semantics in our approach, we fuse the domains in the latent space and generate paths end-to-end to minimize latencies. Moreover, our learned solution goes beyond pure reactiveness by exposing the planner to different behaviors during training while not relying on a globally build map.

\paragraph{Imitation Learning and Self-Supervised Learning}
Another line of research aims to learn local path planning from expert demonstrations~\cite{loquercio_learning_2021, pfeiffer2017perception, zhang2016learning}. For autonomous driving, demonstrations fused with high-definition maps in an end-to-end learned setting allow for applications in environments shared with humans~\cite{hecker2020learning, bao_2023}. In unstructured environments, the traversability estimation module can learn from demonstrations in a self-supervised manner~\cite{caron2021emerging}. In these cases, only weak, platform-dependent supervision is required~\cite{wellhausen_where_2019, frey2023fast}. However, such methods suffer from poor generalization to unfamiliar environments and labor-intense data generation~\cite{bansal2018chauffeurnet}. Additionally, the optimality is limited to the sub-optimality of the expert. On the contrary, we train our method entirely in simulation by simultaneously optimizing the network weights and the task objective to improve generalization while not being bound by sub-optimalities. 

\begin{figure*}[t]
    \centering
    \includegraphics[width=1.0\textwidth]{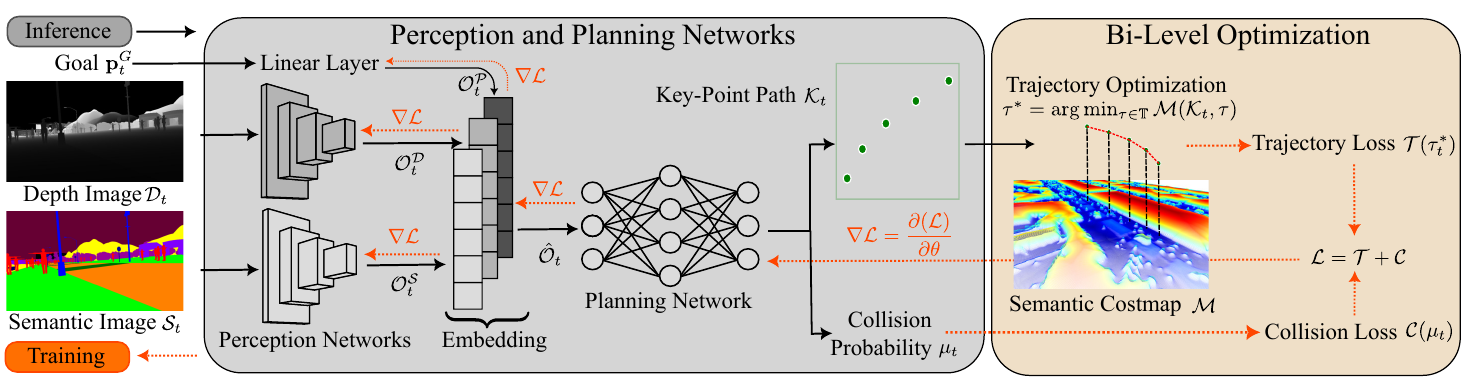}
    \caption{Overview of the integral components of the proposed approach. The perception and planning networks take a depth image, a semantic image, and the desired goal position as input and estimate a coarse plan together with a collision probability. The network weights and final path are jointly optimized as part of a Bi-Level Optimization scheme.}
    \label{fig:network_overview}
\end{figure*}

\paragraph{RL and Imperative Learning}
While previous approaches often relied on extensive expert knowledge and real-world data, Reinforcement Learning (RL) offers a structured approach to end-to-end path planning learning, performed in simulation~\cite{hoeller2021learning}. 
Domain randomization and noised measurements are used to transfer the planner to the real world. However, low sampling efficiency and sparse rewards make training RL models time-intensive, especially for dense input data~\cite{wijmans2019dd}. Most recently, Yang et al.~\cite{yang_iplanner_2023} introduced iPlanner, which uses the Imperative Learning paradigm~\cite{fu2023islam} to address path planning as an offline Bi-Level Optimization. While this concept improves convergence speed compared to RL, iPlanner requires real-world data to bridge the reality gap. Our method builds on top of~\cite{yang_iplanner_2023}, which demonstrated advantages over previous demonstration, modular, and RL-based methods in semi-structured environments, and overcomes its main limitation by going beyond the geometric domain.

\section{Problem Formulation}
\label{sec:problem_formulation}
We define the environment within which the robot operates as $\mathcal{Q} \subset \mathbb{R}^3$. 
Here, $\mathcal{Q}$, consists of non-traversable geometric and semantic obstacles represented by the subset $\mathcal{Q}_{\text{obs}} \subset \mathcal{Q}$, and the traversable space $\mathcal{Q}_{\text{trav}} = \mathcal{Q} \setminus \mathcal{Q}_{\text{obs}}$ where the robot is safe to walk. 
The traversable area $\mathcal{Q}_{\text{trav}}$ is divided into $N$ subsets $\mathcal{Q}_i$ with finite motion costs $c_i \in [0, \inf), \forall i \in \{1,\dots,N\}$.
Consequently, the robot workspace exhibits a fine-grained non-binary separation, fit for representing complex real-world environments, with the complete traversable space $\mathcal{Q}_{\text{trav}}$ defined as $\bigcup_{i=1}^N \mathcal{Q}_i$. The traversability cost of each path $\mathcal{P}$ is defined as its cost integral $\mathcal{T}^\mathcal{T}_\mathcal{P} = \int_\mathcal{P} c(x,y) dp$, with $c(x,y) \in \{c_1,\dots,c_N\}$, depending on the location. The goal cost is defined as the overall length of the path $\mathcal{T}^\mathcal{G}_\mathcal{P} = \int_\mathcal{P}dp$.

\paragraph*{Objective} 
Navigation tasks are ubiquitous in robotics and are generally described as finding a safe, fast, and collision-free path from a start to a goal position in $\mathcal{Q}$.
For this work, we are interested in online motion planning using cheap onboard sensing only. 
In our case, this means that given a depth image observation $\mathcal{D}_t \in \mathbb{R}^{H_D \times W_D}$, a semantic image observation $\mathcal{S}_t \in \mathbb{R}^{H_S \times W_S}$, and an intended goal position $\mathbf{p}_t^G \in \mathcal{Q}_{\text{trav}}$ at timestep $t$ (compare Fig.~\ref{fig:network_overview}), the final goal is to estimate a trajectory $\tau_t = \Phi(\mathcal{D}_t, \mathcal{S}_t, \mathbf{p}_t^G, \theta)$ that guides the robot from its current position $\mathbf{p}_t^R$ to the goal $\mathbf{p}_t^G$, while minimizing the combined traversability and goal cost $\mathcal{T}^\mathcal{T}_\tau + \mathcal{T}^\mathcal{G}_\tau$. Here, $\Phi$ refers to the neural network approximator with weights $\theta$.
Moreover, the collision risk with the environment should be minimized to reduce safety hazards and increase reliability.
It is important to note that this path planning problem must be solved only with partial observations; the two input image streams.


\section{Methodology}
\label{sec:methodology}
The proposed pipeline integrates two stages, as visible in Fig.~\ref{fig:network_overview}.
The first stage consists of the perception and planning networks that encode and concatenate semantic, depth, and goal inputs (Sec.~\ref{sec:semantic_encoding}) and predict a sparse key-point-based path $\mathcal{K}$ towards the goal, along with the collision confidence probability $\mu$ of the generated path (Sec.~\ref{subsec:planning_network}).
The second stage is the BLO process, including the metric-based trajectory optimizer (TO) and network updates. The TO process optimizes the path regarding the semantic costmap (Sec.~\ref{subsec:sem_cost_map}). The TO formulation is introduced in~\cite{yang_iplanner_2023} and will not be discussed in detail in this work for brevity.
The task-level cost function (Sec.~\ref{subsec:train_loss}) to be minimized by both the embedding and planning networks is denoted as $\mathcal{L}$. It consists of trajectory ($\mathcal{T}$) and collision probability ($\mathcal{C}$) costs. 

\subsection{Semantic Encoding}
\label{sec:semantic_encoding}
The presented work utilizes a semantic space of 30 classes typically encountered during navigation challenges. In contrast to other approaches, such as one-hot encoding, our method encodes the traversability directly in RGB colorspace, ensuring that classes with similar traversability are grouped. 
Generally, traversable areas are more within the green, while obstacles are more within the red and blue colorspace.
During training, the deployed neural network then learns to utilize the class characteristics as supplementary information for planning.
Table~\ref{tab:sem_meta_info} provides an overview of all classes with their corresponding color.
\begin{table}[t]
    \centering
    \renewcommand\tabularxcolumn[1]{m{#1}}
    \adjustboxset{height=.01\textheight,
                  valign=c,  margin=0pt 0pt 0pt 0pt}
    \begin{tabularx}{\columnwidth}{Y c p{2cm}}
    \toprule
    \textbf{Classes} & \textbf{Cost} & \textbf{Spectrum}\ \\
    \hline\hline
    
    sidewalk, crosswalk, floor, stairs & $c_{free}$ & \adjustimage{width=2cm}{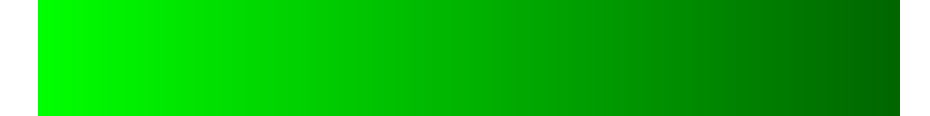}\\
    \hline
    
    gravel, sand, snow & $c_{mid_1}$ & \adjustimage{width=2cm}{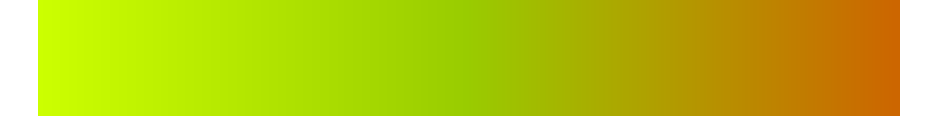}\\
    \hline
    
    terrain (grass, dirt) & $c_{mid_2}$ & \adjustimage{width=2cm}{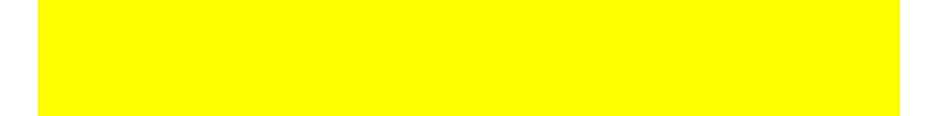}\\
    \hline
    
    road & $c_{mid_3}$ & \adjustimage{width=2cm}{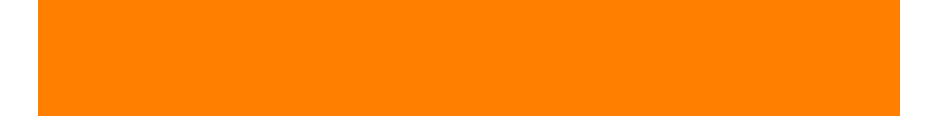}\\
    \hline
    
    person, animal, vehicle, trains, motorcycle, bicycle & $c_{obs}$ & \adjustimage{width=2cm}{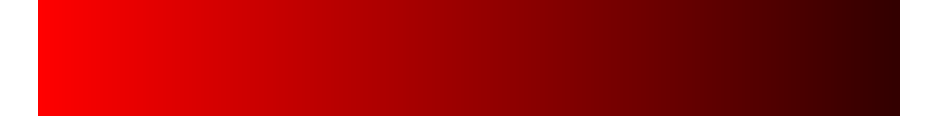}\\
    \hline
    building, wall, fence, bridge, tunnel, furniture, tree, water surface& $c_{obs}$ & \adjustimage{width=2cm}{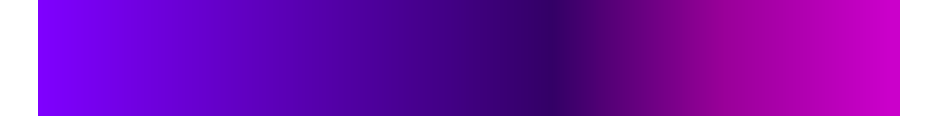}\\
    \hline
    pole, traffic sign, traffic light, bench & $c_{obs}$ & \adjustimage{width=2cm}{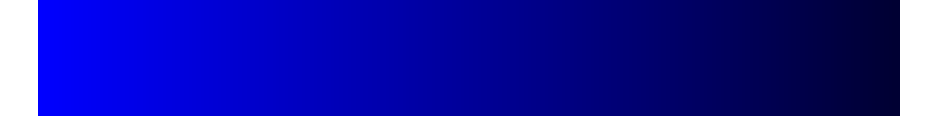} \\
    \hline
    sky, ceiling, unknown & $c_{obs}$ & \adjustimage{width=2cm}{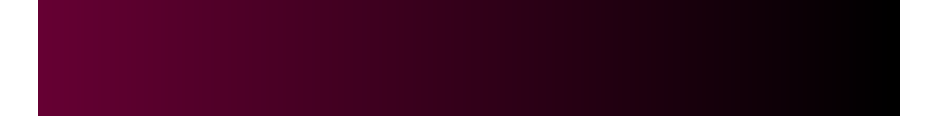} \\

    \bottomrule
    \end{tabularx}
    \caption{RGB-encoded semantic colorspace for navigation. While each class has its color, in this table, multiple similar classes are grouped by spectrum. Each group is associated with a motion cost~$c$, with $c_{free}$ the lowest and $c_{obs}$ the largest.}
    \label{tab:sem_meta_info}
\end{table}

\subsection{Perception and Planning Networks}
\label{subsec:planning_network}

\subsubsection{Perception Networks}
For every time stamp $t$, the two perception networks receive a depth and semantic image, respectively. At their core, both networks consist of a ResNet-18~\cite{he2016deep} architecture and are trained from scratch without weight sharing. During inference, a separate semantic segmentation network generates $\mathcal{S}_t$ from raw RGB input. 
We transform both measurements into the same camera frame and estimate embeddings $\mathcal{O}^{\mathcal{D}} \in \mathbb{R}^{C_I \times M}$ and $\mathcal{O}^{\mathcal{S}} \in \mathbb{R}^{C_I \times M}$.
\subsubsection{Combined Feature Embedding}
To retrieve the target efficiently, the commanded goal position $\mathbf{p}_t^R$ is mapped by a linear layer to a higher dimensional embedding $\mathcal{O}^{\mathcal{P}}_t \in \mathbb{R}^{C_G \times M}$, with $C_G \geq 3$.
We then concatenate the embeddings $\mathcal{O}^{\mathcal{D}}_t$ and $\mathcal{O}^{\mathcal{S}}_t$ with this goal position embedding, resulting in the combined embedding $\hat{\mathcal{O}}_t \in \mathbb{R}^{(2C_I + C_G) \times M}$. 
This embedding is essential, as it constitutes the input to the planning network.
\subsubsection{Planning Network}
Our lightweight planning network consists of convolutional layers (CNN) and a multilayer perceptron (MLP). Building on~\cite{yang_iplanner_2023}, it includes two distinct heads: the \textit{path planning}- and the \textit{collision probability} head.
The former predicts a sparse set of key points $\mathcal{K}_t \in \mathbb{R}^{n_k \times 3}$ that are the core input to the trajectory optimization.
The latter estimates the risk of collision $\mu$ with obstacles for each trajectory and acts as a supplementary safety measure. 
This estimate is necessary, as a naive increase of the loss for obstacle violations leads to overly conservative policies. 
Instead, the collision head allows for more flexibility, which is crucial in scenarios where the system is trapped in local minima.
Only trajectories with a collision probability of less than $\delta_{\mu}=0.5$ are executed during online inference.

\subsection{Semantic Costmap}
\label{subsec:sem_cost_map}
\begin{figure}[t]
    \centering
    \begin{tikzonimage}[width=0.49\columnwidth]{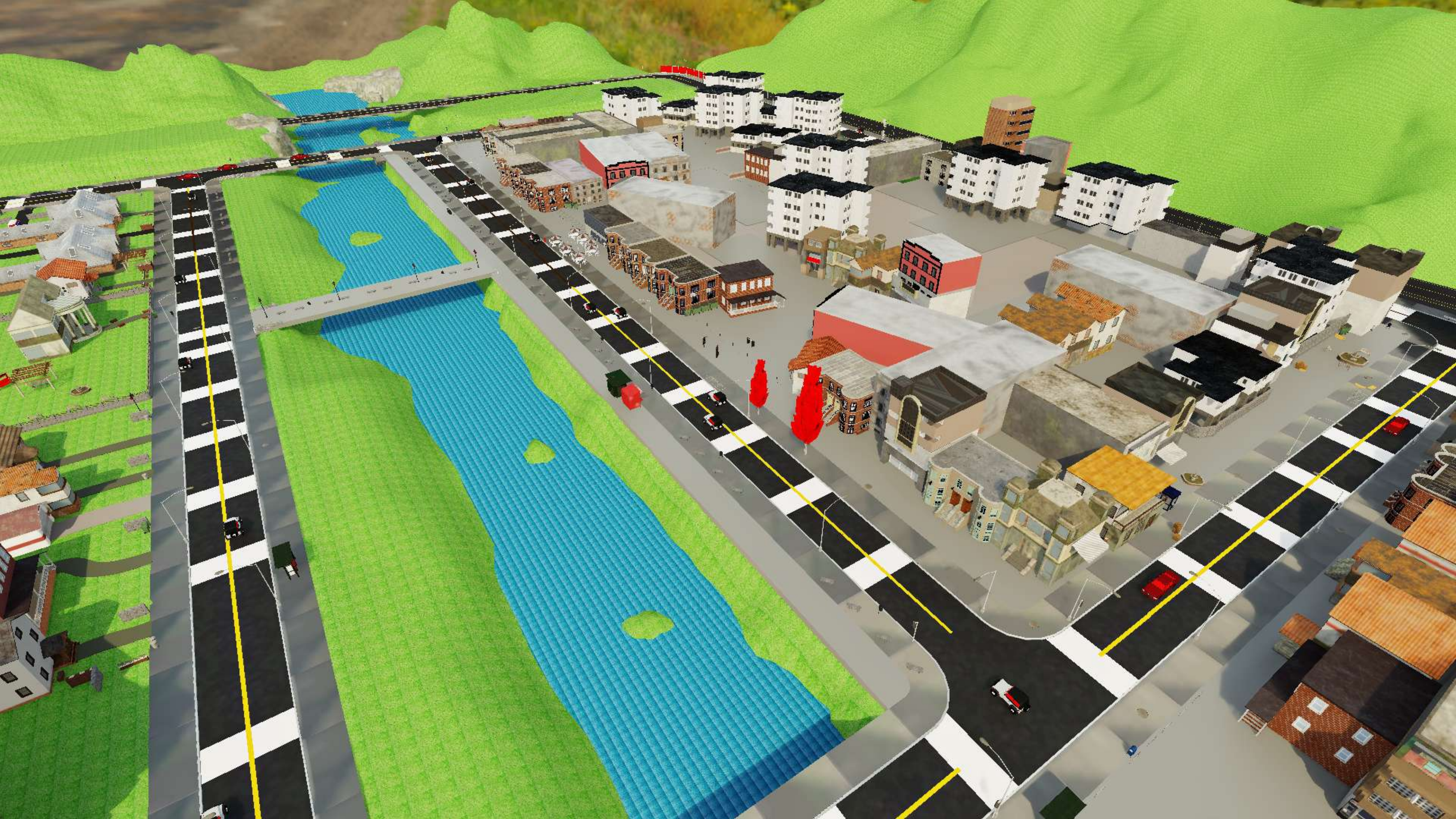}[image label]
        \node{RGB};
    \end{tikzonimage}
    \begin{tikzonimage}[width=0.49\columnwidth]{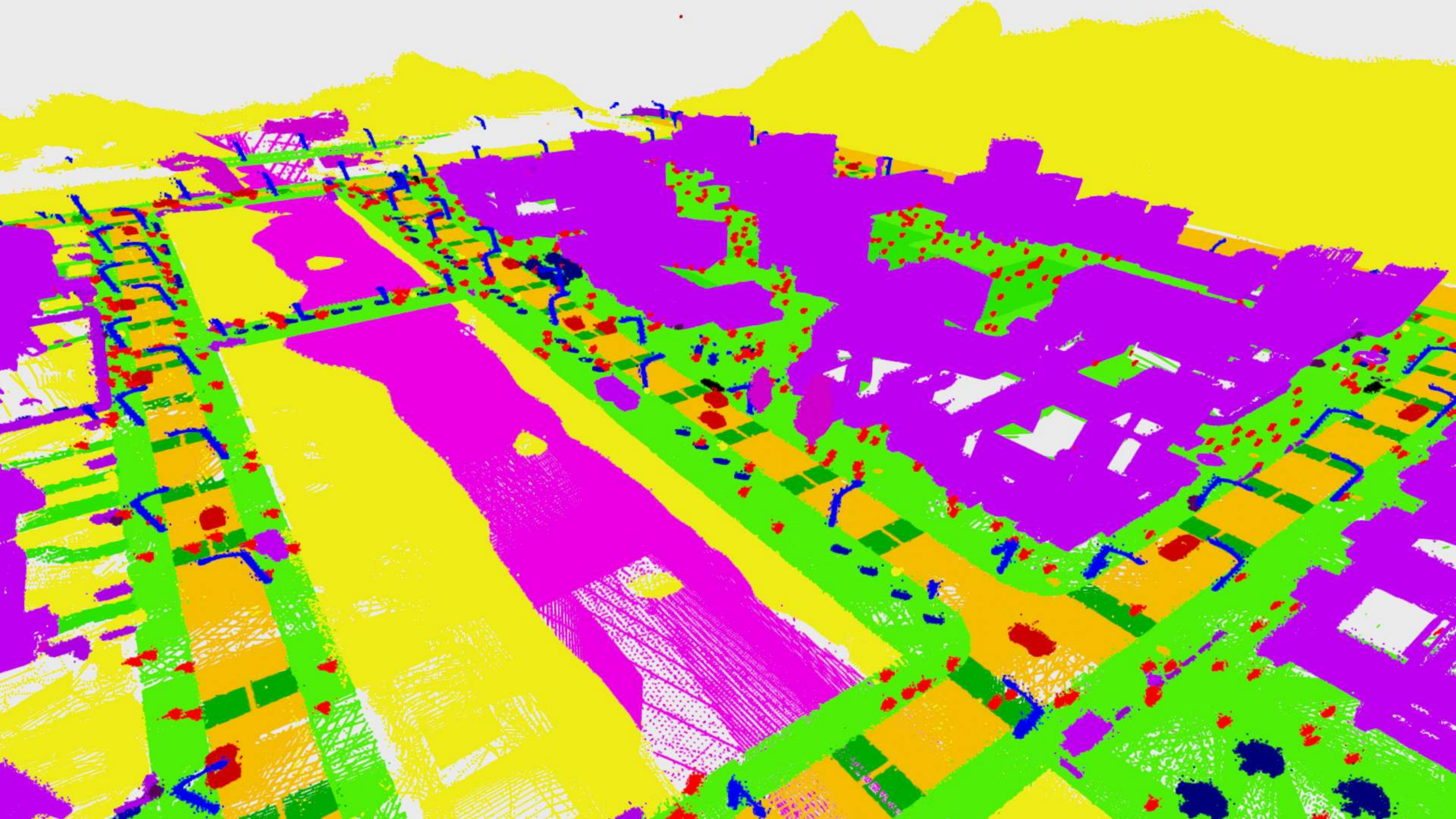}[image label]
        \node{Reconstruction};
    \end{tikzonimage}
    \begin{tikzonimage}[width=0.49\columnwidth]{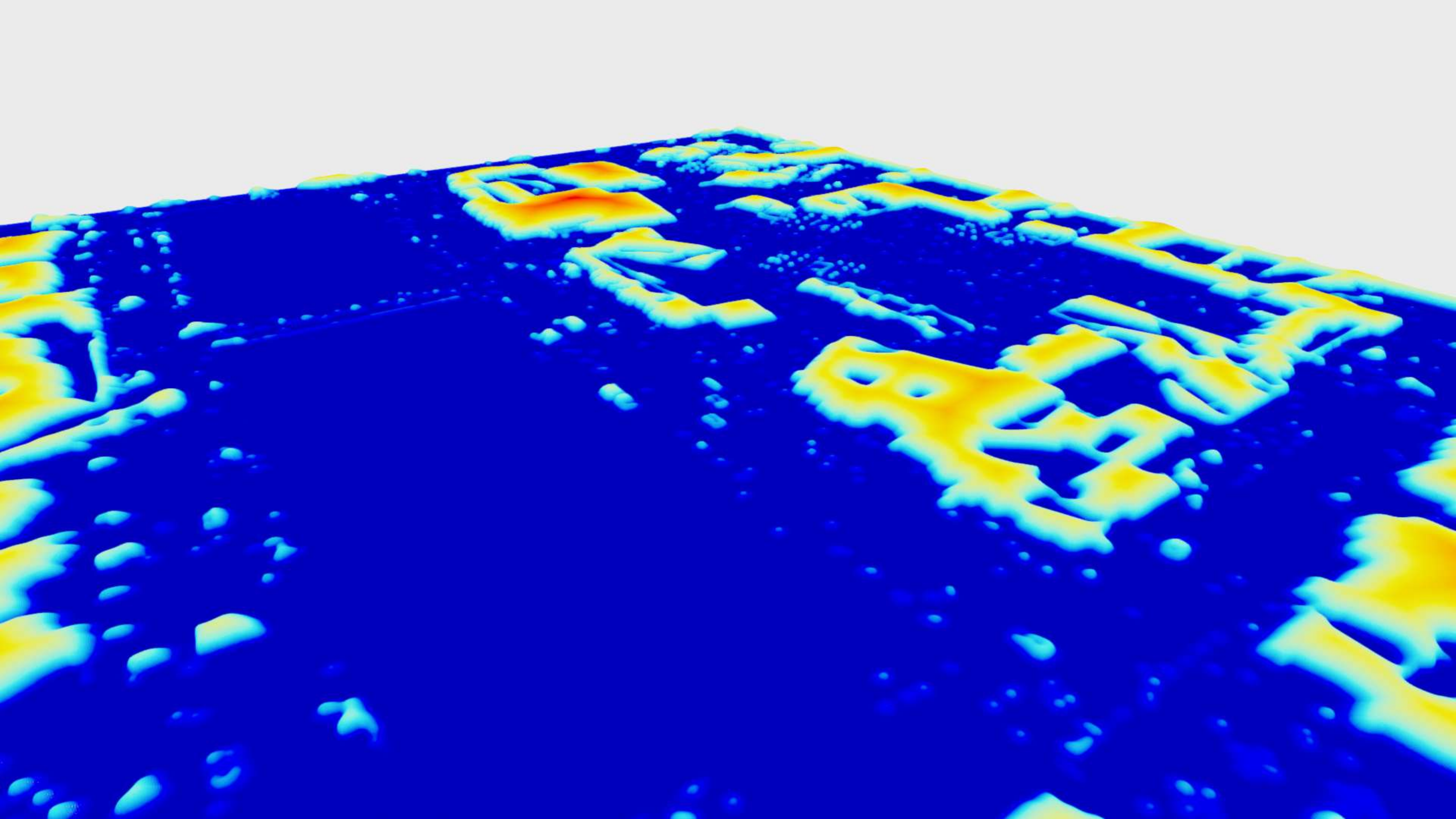}[image label]
        \node{Geometric Cost};
    \end{tikzonimage}
    \begin{tikzonimage}[width=0.49\columnwidth]{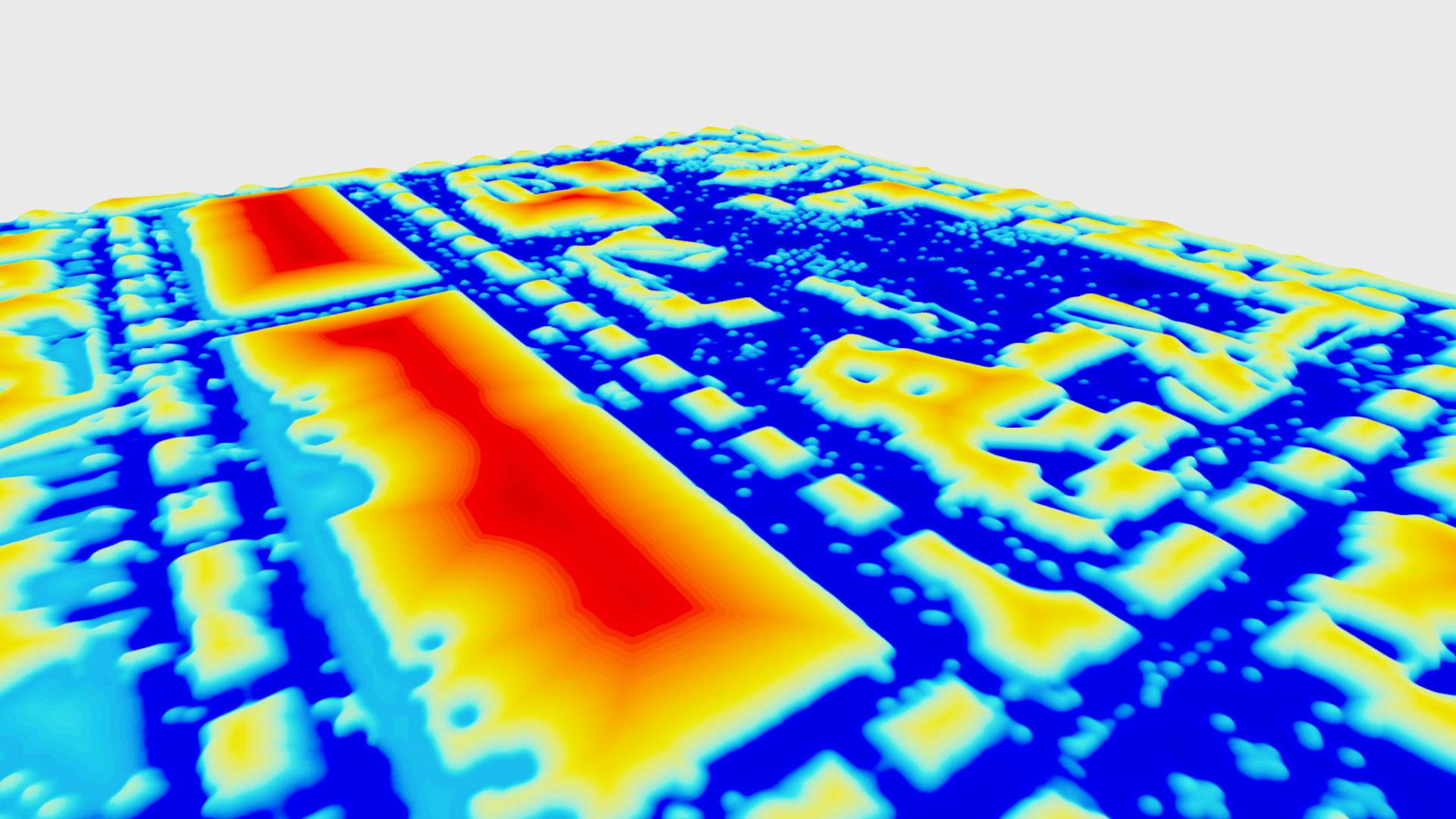}[image label]
        \node{Semantic Cost};
    \end{tikzonimage}
\caption{Example training environment for the urban CARLA dataset~\cite{dosovitskiy2017carla} with its semantic reconstruction, and created geometric and semantic costmaps.}
\label{fig:cost_map}
\end{figure}

The creation of the semantic costmap~$\mathcal{M}$ and its usage during the trajectory and network optimization is one of the core components of this work. Note, however, that this costmap is only needed during training time. During inference, the condensed policy estimates the resulting paths from the incoming image stream.

To obtain the costmap, we create a 2D grid with a set resolution of the size of the environment. 
Each cell is assigned a class label depending on the mesh at the corresponding location. Fig.~\ref{fig:cost_map} shows the semantically annotated environment as a point-cloud reconstruction. 
Cost factors, as given in Tab.~\ref{tab:sem_meta_info}, are assigned to each cell based on the class. 
We require the costmap to be differentiable and smooth to enable successful (network) optimization. 
The necessary smoothing is carried out by first applying a Gaussian filter to remove possible classification errors without affecting small obstacles and second by using a signed-distance gradient value toward the closest class boundaries, to reduce the impact of areas with constant loss. 
For the area with the smallest loss, this operation is inverted with a gradient pointing towards the center, guiding the robot to the center of elongated parts of the map (such as hallways). 
This step is crucial to accelerate the training and push the path towards the lowest loss areas. 
Moreover, it enables us to effectively train in outdoor environments with large spaces of constant loss values.
Lastly, a second Gaussian Filter is applied to smooth the boundaries between areas of different classes. An example of a resulting costmap is shown in Fig.~\ref{fig:cost_map}.

\subsection{Training Loss}
\label{subsec:train_loss}
As introduced in~\cite{yang_iplanner_2023}, the ImpL training loss $\mathcal{L}(\tau_t)$ includes a trajectory loss term, $\mathcal{T}_t(\tau_t)$ and a collision loss term, $\mathcal{C}_t(\tau_t, \mu_t)$. We directly adopt the definition of $\mathcal{C}(\tau, \mu)$, 
\begin{equation}
    \mathcal{C} = 
    \begin{cases}
        \text{BCELoss}(\mu, 0.0)   & \mathbf{p_i}^K \in \mathcal{Q}_{tra} \forall \mathbf{p_i}^K \in \tau \\
        \text{BCELoss}(\mu, 1.0)   & \text{otherwise},
    \end{cases}
\end{equation}
with the Binary-Cross-Entropy Loss (BCELoss) and the center points $\mathbf{p}_i^K$ of the path $\tau$. Our implementation of $\mathcal{T}$ differs from the original formulation in multiple ways.
The resulting trajectory loss term is formulated as follows:
\begin{equation}
    \mathcal{T}(\tau) = \alpha \mathcal{T}^{\mathcal{T}} + \beta \mathcal{T}^{\mathcal{G}} + \gamma \mathcal{T}^{\mathcal{M}} + \delta \mathcal{T}^{\mathcal{H}}.    
\end{equation}
Here, $\alpha, \beta, \gamma$ and $\delta$ are  for loss scaling, and $\mathcal{T}^{\mathcal{M}}$ is the same motion loss term as used in~\cite{yang_iplanner_2023}.
Our work introduces new formulations for the obstacle, now called \textit{traversability} $(\mathcal{T}^{\mathcal{T}})$ and goal $(\mathcal{T}^{\mathcal{G}})$ terms, and extends the formulation with an additional height loss $\mathcal{T}^{\mathcal{H}}$. 

\subsubsection{Traversability Loss}
\label{subsubsec:traversability_loss}
The new traversability loss takes the physical size of the robot into account by evaluating the cost not only at the center points $\mathbf{p}_i^K$, but also at points perpendicular to the path at a distance $w^R$, corresponding to the robot width.
The resulting loss is
\begin{equation}
\label{equation:obstacle_loss}
\!\begin{aligned}
    & \mathcal{T}^{\mathcal{T}}(\tau) = \frac{1}{3n} \sum_{i=1}^{n} \Tilde{m}(\mathbf{p}^K_i) + \Tilde{m}(\mathbf{p}^K_i + w^R \cdot \mathbf{n}^K_i) \\
    & \qquad \qquad \qquad \qquad + \Tilde{m}(\mathbf{p}^K_i - w^R \cdot \mathbf{n}^K_i),    
\end{aligned}
\end{equation}
where $\Tilde{m}(\cdot)$ is the bi-linearly interpolated value of the costmap $\mathcal{M}$ at the corresponding point. 

\subsubsection{Goal Loss}
For the goal loss, we introduce a log scaling to limit the influence of goal points at large distances. The loss can be expressed as
\begin{equation}
    \mathcal{T}^{\mathcal{G}}(\tau) = \log(\| \mathbf{p}_{n}^K - \mathbf{p}_i^G \|_2 + 1.0),
\end{equation}
with $\mathbf{p}_{n}^K$ denoting the last point of the path. 

\begin{figure*}
    \begin{tikzonimage}[width=0.33\textwidth]{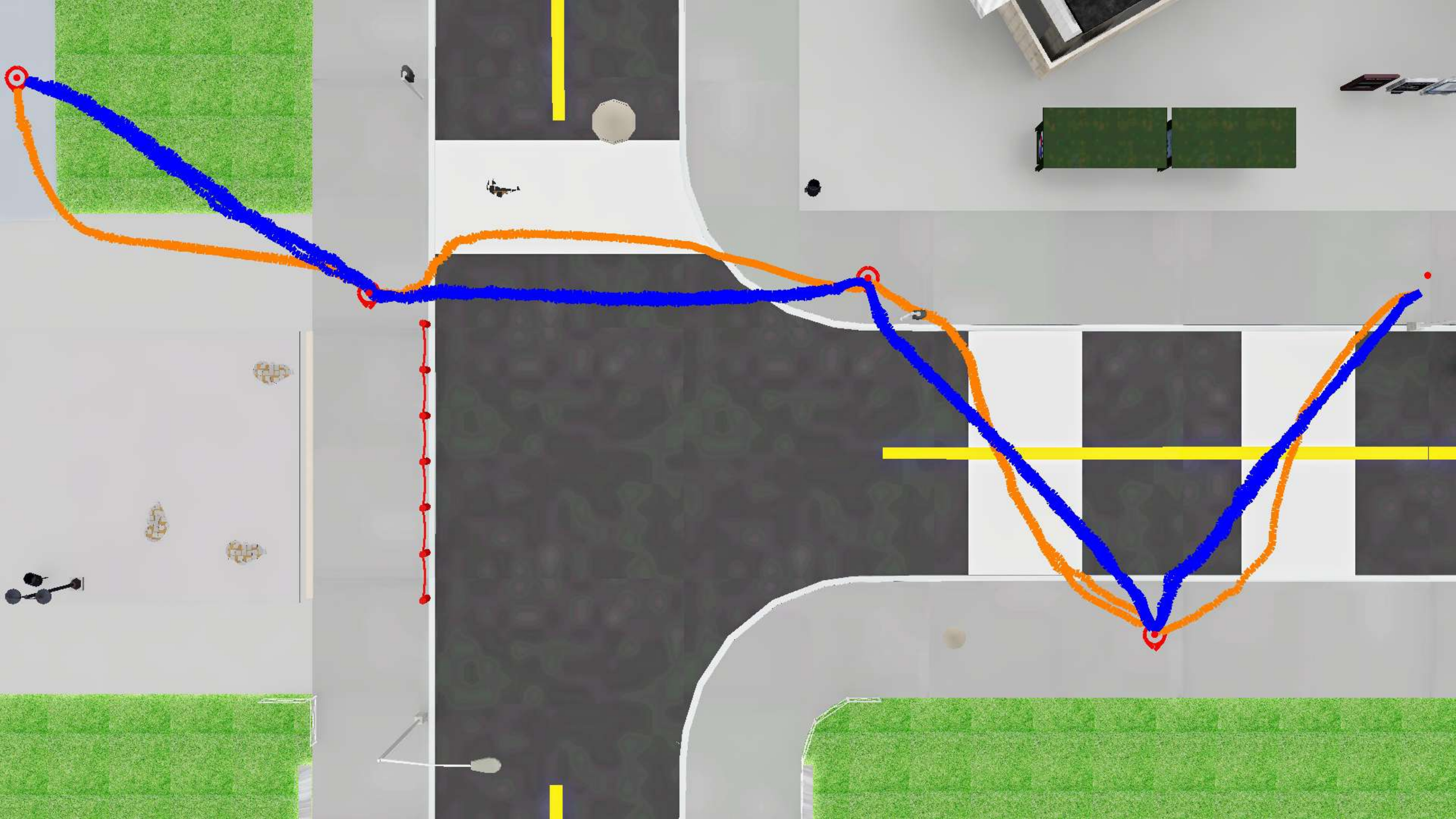}[image label_bl]
        \node{(a) CARLA};
    \end{tikzonimage}
    \begin{tikzonimage}[width=0.33\textwidth]{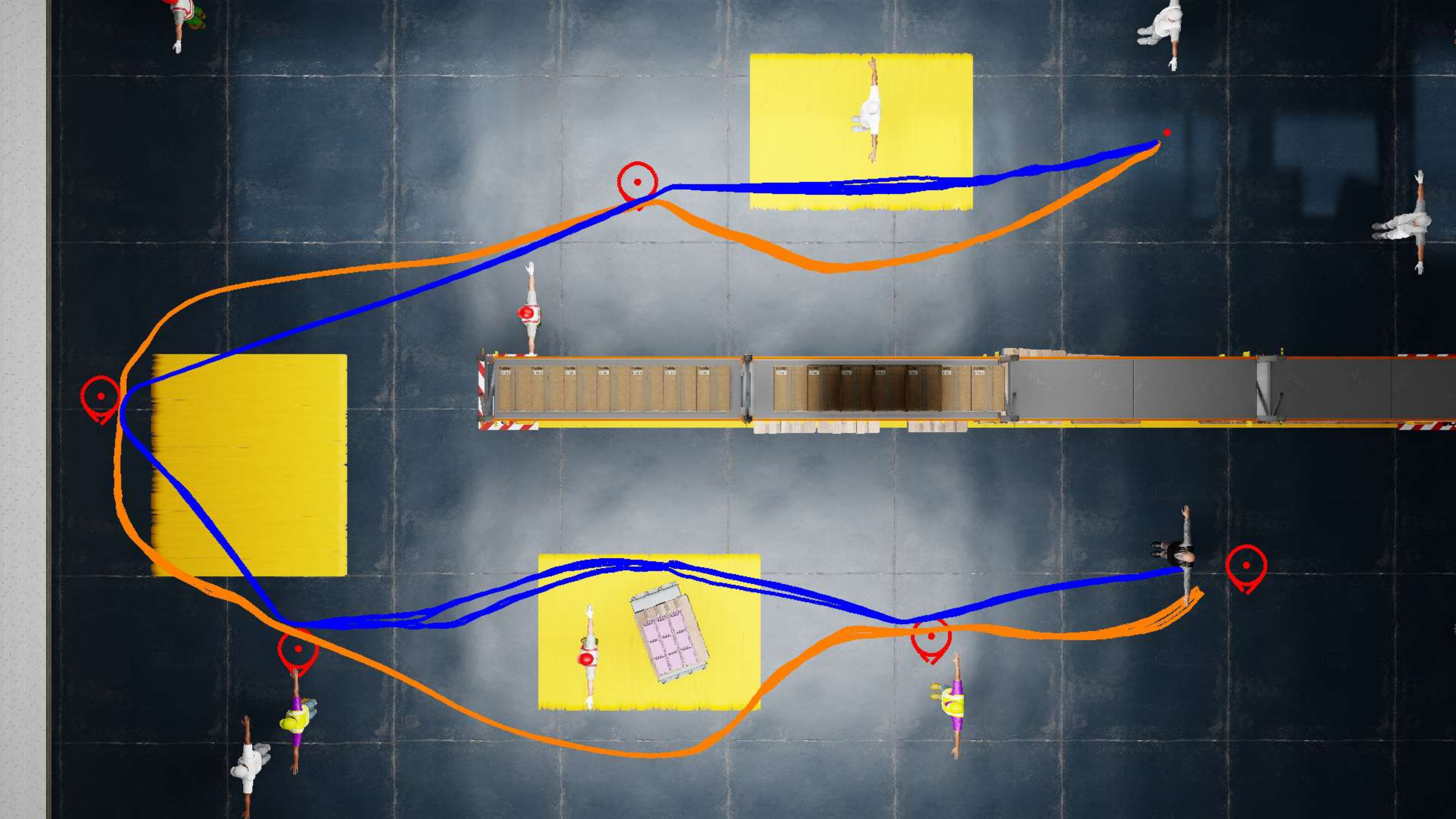}[image label_bl]
        \node{(b) Warehouse};
    \end{tikzonimage}
    \begin{tikzonimage}[width=0.33\textwidth]{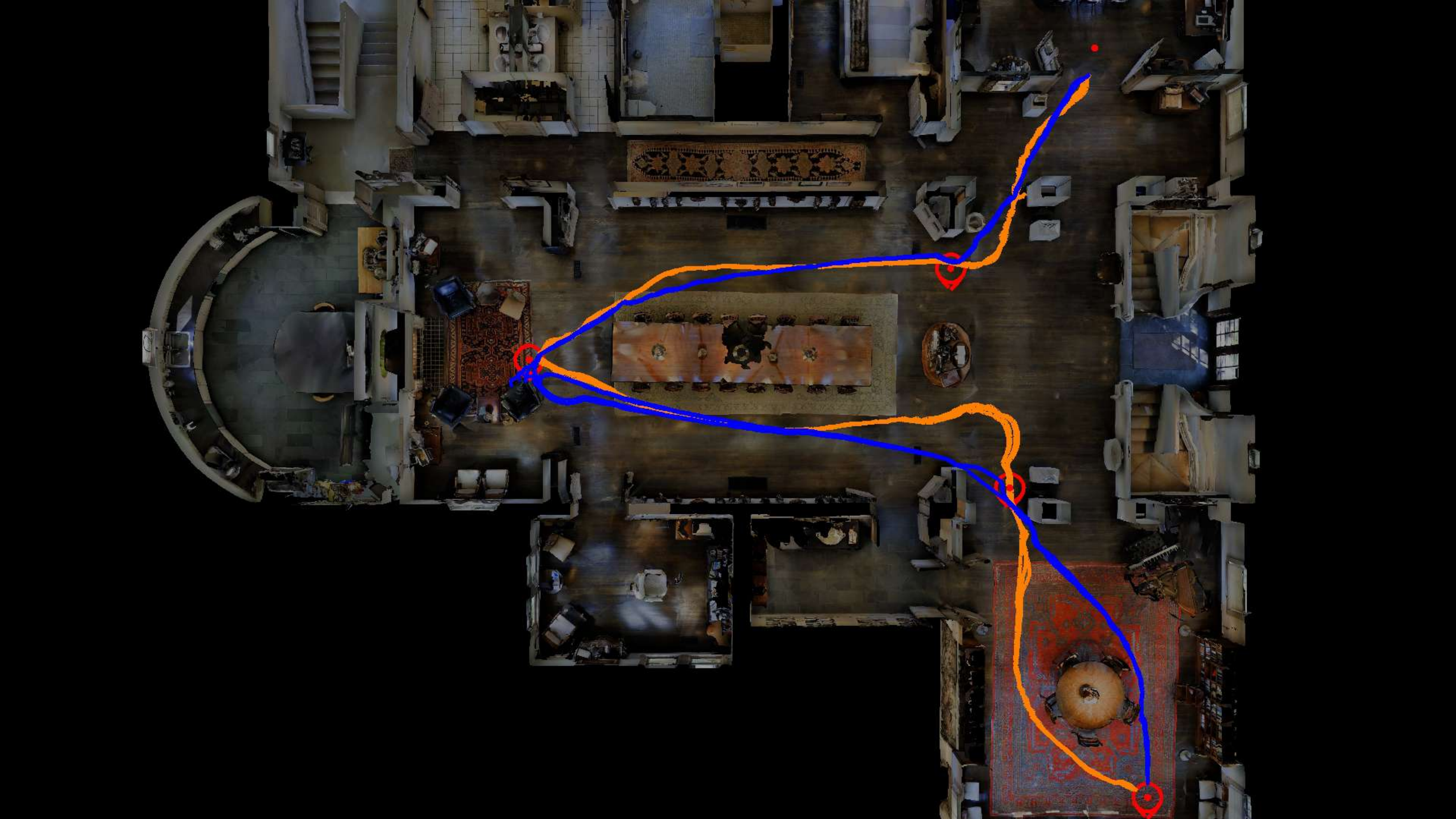}[image label_bl]
        \node{(c) Matterport};
    \end{tikzonimage}
    \caption{Qualitative comparison between the proposed ViPlanner (orange) and the purely geometric iPlanner~\cite{yang_iplanner_2023} (blue). Five trials of both planners are shown for the manually selected waypoints (red). The street (a) and the yellow working areas (b) are successfully avoided.}
    \label{fig:sim2sim}
\end{figure*}

\subsubsection{Height Loss}
The additional, newly introduced height loss regularizes the path to maintain the base height of the robot $h^R$, avoiding any maneuvers that might circumvent obstacles by going above or below them. The corresponding loss is given as
\begin{equation}
    \mathcal{T}^{\mathcal{H}}(\tau_t) = \frac{1}{n} \sum_{i=1}^n \vert z_{p, i}^K - \Tilde{h}(\mathbf{p}_i^K) - h^R \vert,
\end{equation}
where $\Tilde{h}(\cdot)$ is the bilinearly interpolated value of the height-map $\mathcal{H}$ at the given point.


\section{Implementation}
\label{sec:implementation}

\paragraph*{Simulation Environments} 
The proposed planner has been fully trained with \textit{NVIDIA Omniverse} and evaluated using the legged robot ANYmal~\cite{hutter2017anymal} and a pre-trained RL locomotion policy provided in the \textit{Orbit} Framework~\cite{mittal2023orbit}. 
To allow for successful navigation in semi-structured environments, we picked realistic indoor scenes from the Matterport3D dataset~\cite{chang2017matterport3d} and relevant outdoor scenes as released in CARLA~\cite{dosovitskiy2017carla}, compare Fig.~\ref{fig:sim2sim}.
We developed new plugins to load both datasets into \textit{Omniverse}, to benefit from its scalability and realistic physics engine. This allows us to simulate the full robot model, including physical contacts, instead of a perfect point approximation for evaluation, as done in~\cite{yang_iplanner_2023}.
All code, including the dataset processing and necessary plugins, are made publicly available and will help develop and test future applications.

\paragraph*{Training Set Creation}
For the successful deployment of our method and scaling to larger amounts of data, a flexible and automatic sampling procedure for capturing diverse (semantic) image viewpoints within the environments is required.
In contrast to the works in~\cite{nubert2022learning, yang_iplanner_2023}, where a similar problem has been solved through random-pose sampling around manually defined paths, in this work, the sensor model is spawned fully automatically in simulation according to the so-called \textit{Halton} sequence~\cite{Halton1960} at robot-accessible locations. 
The corresponding images at each sampled point are then rendered at angles that maximize the coverage of the traversable space. 
Moreover, goal points are placed randomly at the sampled camera centers from before. 
Their reachability is identified by constructing a graph between all centers and removing connections that pass through areas on the costmap~$\mathcal{M}$ higher than a certain threshold. 
For successful obstacle avoidance learning, the goal should preferably be within the robot's field of view (FoV). Thus, our data generation pipeline allows us to define the desired ratio of the relative goal point locations w.r.t. the FoV.
Overall, we generated approx. $80k$ start-goal pairs from eleven Matterport (avg. $36 \times 33$m), one CARLA ($400 \times 400$m) and three warehouse (avg. $25 \times 35$m) environments.


\section{Experiments}
\label{sec:experiments}

\paragraph*{Model Training}
The training process is managed using the SGD optimizer, a learning rate scheduler, and an early-stopping strategy. The training procedure is executed on an \textit{NVIDIA RTX 3090} for around six hours. The semantic cost factors, compare Tab.~\ref{tab:sem_meta_info}, are in the range of zero to two.

\paragraph*{Experiment Setup}
We demonstrate the effectiveness of our model, data generation, and training strategy by comparing it in various simulation environments to~\cite{yang_iplanner_2023}. In addition, we employ our method on the legged robot ANYmal~\cite{hutter2017anymal} to showcase the zero-shot transfer to the real world. The planner runs on a \textit{Nvidia Jetson Orin AGX} on the robot and uses the latest semantic image generated by the state-of-the-art \textit{Mask2Former} segmentation model~\cite{cheng2022masked} and the current depth image as input. The semantic segmentation model has been fine-tuned on a small set of images collected in Zurich, Switzerland, where we conducted our outdoor experiments. By asynchronously reusing the semantic images generated at approximated 3Hz, an independent planner frequency of 10Hz can be achieved.

\begin{figure*}[t]
    \centering
    \includegraphics[width=0.95\textwidth]{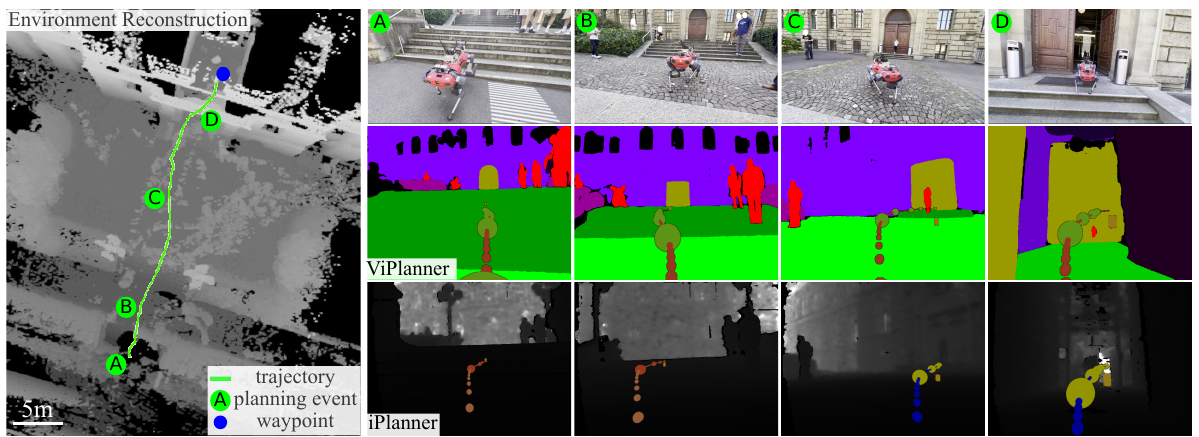}
    \caption{Real-world experiment in the presence of geometric occurrences such as stairs. Four planning events \mbox{(A - D)} along the path with corresponding semantic and depth images are shown with the estimated paths of the proposed planner and of iPlanner. iPlanner's first two predictions are marked completely in red (indicating a high estimated collision probability of $\sim0.98>\delta_{\mu}$),  causing the robot to stop.}
    \label{fig:real_world_stairs}
\end{figure*}

\subsection{Simulation Experiments}

\begin{table}[b]
    \centering
    \renewcommand{\tabcolsep}{3pt}
    \scriptsize
    \begin{tabular}{l | c | *{2}{c} | c}
        \toprule
        Scene & Method & Geom. Loss & Sem. Loss & Goal Reached \\
        \midrule
        \multirow{3}{*}{Matterport3D} & iPlanner & $1.77 \pm 0.379$ & $1.29 \pm 0.376$ & $55.37$\% \\  
         & Ours (Geom) & $1.76 \pm 0.543$ & $1.27 \pm 0.538$ & $71.05$\%\\
         & Ours (Sem) & $\mathbf{1.70 \pm 0.412}$ & $\mathbf{1.19 \pm 0.410}$ & $\mathbf{71.60\%}$\\
        \midrule
        \multirow{3}{*}{CARLA} & iPlanner & $1.15 \pm 0.653$ & $1.17 \pm 0.669$ & $79.59$\% \\  
         & Ours (Geom) & $1.03 \pm 0.579$ & $1.06 \pm 0.708$  & $\mathbf{95.91}$\% \\
         & Ours (Sem) & $\mathbf{1.00 \pm 0.602}$ & $\mathbf{0.69 \pm 0.530}$  & $87.75$\% \\
        \midrule
        \multirow{3}{*}{Warehouse} & iPlanner & $0.90 \pm 0.544$ & $1.793 \pm 1.274$  & $86.56$\% \\  
         & Ours (Geom) & $0.82 \pm 0.436$ & $1.641 \pm 1.283$  & $\mathbf{88.72}$\% \\
         & Ours (Sem) & $\mathbf{0.71 \pm 0.396}$ & $\mathbf{0.96 \pm 1.041}$  & $85.58$\%\\
        \bottomrule
    \end{tabular}  
    \caption{Comparison between the geometric iPlanner~\cite{yang_iplanner_2023} against our method trained with geometric data only (geom) and with geometric and semantic information (sem). 
    Geom. and Sem. Loss metrics refer to the traversability loss of the trajectories evaluated on the geometric or semantic cost map respectively.
    Adding the semantics leads to recognizing geometrically invisible obstacles and interrupting more paths due to collision probabilities larger than $\delta_{\mu}$.}
    \label{tab:sim_metric_subsequent}
\end{table}

We test our proposed planner in three diverse simulation environments: i) the \textit{CARLA} urban dataset, ii) the \textit{NVIDIA} warehouse environment, iii) the indoor \textit{Matterport3d} dataset (Fig.~\ref{fig:sim2sim}), and compare it against the previous geometric iPlanner~\cite{yang_iplanner_2023}. 
Both methods are trained from scratch with the data generated in our proposed pipeline.
Three navigation tasks are shown in Fig.~\ref{fig:sim2sim}, for which the predictions of the two planners are provided. Intuitively, the proposed planner successfully uses both semantic and depth information to traverse low-cost areas, such as crosswalks in the urban environment, and avoid geometric and semantic obstacles, such as shelves or yellow working areas (treated as class unknown) in the warehouse environment. 
In contrast, the geometric planner perceives no difference between the terrains, resulting in straight motions towards the goal position.

Next, we evaluate our proposed method quantitatively by running experiments on a larger scale. We train our planner once with access only to the geometric domain and once with the fusion of depth and semantics to highlight the effects of the new loss and costmap design. 
Table~\ref{tab:sim_metric_subsequent} shows the results of $500$ random paths with unique start-goal configurations in each of the three simulated environments. We report the rate of paths that reach the goal up to a threshold distance of $0.5m$ and the traversability loss based on geometric and semantic costmaps for these paths.
When comparing iPlanner to ViPlanner trained only with the geometric domain, a constant decrease over both costmaps is evident, together with an average increase of 16\% in reached goal points for the Matterport3D and Carla environment. 
Due to the spacious layout of the warehouse environment, paths are easier to reach, which results in a less significant decrease.
This loss improvement can be attributed to the usage of the robot width in Equation~\eqref{equation:obstacle_loss}, compared to the point approximation of the robot used in iPlanner and the different smoothing process of the costmap.
Our planner succeeds in using the semantic information, evident in an average semantic loss decrease of 38.02\% in the CARLA and warehouse environment, where not all obstacles are detectable through geometry, e.g., the roads.
In the same environments, a more conservative behavior of the proposed method can be observed, as the planner is aware of high-cost areas in the semantic domain, such as a road or a working area. Due to the unawareness of the geometric planners, they are not bound by these additional constraints and proceed to the goal, resulting in a higher percentage of "Goal Reached".

\subsection{Real-World Experiments}
We demonstrate zero-shot sim-to-real capabilities in two experiments, contrasting previous work such as~\cite{yang_iplanner_2023,hoeller2021learning}, where real data must be mixed into the training process.
Our planner, trained solely on simulated data, 
has been employed in various scenarios despite diverse lighting conditions, new obstacle shapes, unknown scene compositions, and sensor noise.

The first experiment is shown in Fig.~\ref{fig:real_world_crosswalk}, for which the planner is tasked to navigate in an urban outdoor environment. As the goal is to navigate safely, the planner correctly commands the robot to cross the street at the crosswalk location and proceed further on the sidewalk. 
With the provided sensor measurements, it recognizes lower cost areas and demonstrates adjusted paths that consider the geometrically invisible constraints.
Larger distances can be traversed with a small number of manually selected waypoints. 
The second experiment investigates the planners's ability to handle seemingly geometric obstacles that, in reality, can be overcome by a legged robot, such as stairs.
Fig.~\ref{fig:real_world_stairs} showcases parts of the experiment, highlighting the ability to recognize and navigate stairs correctly.
A comparison to iPlanner shows that a purely geometric approach struggles in such situations, and its collision estimates ($> \delta_\mu$) will prevent the execution of the predictions, ultimately failing to reach the goal position.

\section{Conclusions \& Future Work}
\label{sec:conclusion}

In this work, we presented a semantic-aware end-to-end trained local planner for deployment in semi-structured environments, fully trained on simulated data and transferable to the real world. 
With the presented loss and costmap design, up to 16\% more goals can be reached compared to previous geometric approaches. 
Moreover, by fusing geometric and semantic information, our planner decreases the semantic traversability loss by up to 42\%. The developed simulation plugins, the scalable data generation pipeline, and the planner models are open-sourced to expedite forthcoming research. 

For future work, we will investigate how to remove hand-crafted loss values to make the costmap more generic and work with open-vocabulary representations.
In addition, we aim to determine the sensitivity against semantic segmentation errors.
Also, we will incorporate memory to enhance consistency and prevent the planner from "forgetting" obstacles.

\section*{Acknowledgement}
The authors thank Turcan Tuna and David H\"{o}ller for their support during experiments and the scientific discussions.


\bibliographystyle{IEEEtran}
\bibliography{bibliography/references}

\end{document}